\documentclass[11pt,a4paper]{article}
\usepackage[hyperref]{emnlp2020}
\usepackage{amsmath}
\usepackage{amssymb}
\usepackage[most]{tcolorbox}
\usepackage{times}
\usepackage{latexsym}
\usepackage{textcomp}

\usepackage[nameinlink,capitalise]{cleveref}
\usepackage{booktabs}
\usepackage{dirtytalk}
\usepackage{natbib}
\usepackage{url}
\usepackage{float}

\Crefname{figure}{Figure}{Figures}

\DeclareMathOperator*{\MVN}{\textrm{MVN}}
\DeclareMathOperator*{\unif}{\textrm{unif}}

\aclfinalcopy 

\title{Is the Best Better? Bayesian Statistical Model  \\
Comparison for Natural Language Processing}

\author{Piotr Szymański \\
  Department of Computational Intelligence \\
  Wrocław University of Science and Technology \\
  {\tt piotr.szymanski@pwr.edu.pl} \\ \And
  Kyle Gorman \\
  Graduate Program in Linguistics \\
  City University of New York \\
  {\tt  kgorman@gc.cuny.edu} \\}
\date{}

\begin{document}
\maketitle
\begin{abstract}
Recent work raises concerns about the use of standard splits to compare natural language processing models.
We propose a Bayesian statistical model comparison technique 
which uses $k$-fold cross-validation across multiple data sets to estimate the likelihood that one model will outperform the other, or that the two will produce practically equivalent results.
We use this technique to rank six English part-of-speech taggers across two data sets and three evaluation metrics.
\end{abstract}

\section{Introduction}

\citet{gorman-bedrick-2019-need} raise concerns about standard procedures used to compare speech and language processing models.
They evaluate the performance of six English part-of-speech taggers
using multiple randomly-generated training-testing splits;
in some cases, they fail to reproduce previously-published system rankings established using a single \say{standard} split.
They argue that point estimates of performance derived from a single training-testing splits are insufficient to establish system rankings,
even when null hypothesis significance testing is used for model comparison.

In this study, we propose a technique for system comparison based on Bayesian statistical analysis.
Our approach,
motivated in \cref{sec:prior-work}
and
described in \cref{sec:model-comparison},
allow us to infer the likelihood that one model will outperform the other,
or even that both models' performance will be practically equivalent, something that is not possible with the frequentist statistical tests used by \citeauthor{gorman-bedrick-2019-need}.
Our approach can also be applied simultaneously across multiple data sets.
As a proof of concept, in \crefrange{sec:methods}{sec:results} we apply the proposed method using the experimental setup of \citeauthor{gorman-bedrick-2019-need},
and in \cref{sec:discussion}, we use it to rank the six taggers, compare evaluation metrics, and interpret the results.
Our failure to reproduce some of earlier reported results leads us to discuss the impact of repeating experiments, contrasting performance in multiple measures and the advantages of comparing likelihoods in \cref{sec:discussion}. We also discuss the notion of \emph{practical equivalence} for speech and language technology.

\section{Prior work}
\label{sec:prior-work}

\citet{langley1988machine} argues that machine learning should be viewed as an experimental science,
and as such, machine learning technologies should be evaluated according to their performance on multiple held-out data sets.
\citet{Dietterich98} proposes a framework for comparing two supervised classifiers using a null hypothesis tests to determine whether two classifiers have the same likelihood of predicting a correct result.
This study introduces several methods, including a paired $t$-test for $k$-fold cross-validation results.
However, he notes that the assumptions of normality and independence may not be satisfied in all cases.
\citet{nadeau2000inference} propose a correlation-based correction for the \citeauthor{Dietterich98} $t$-test procedure which adjusts for the overlap between folds.
\citet{hull1994information} and \citet{schutze1995comparison} 
propose non-parametric tests for comparing models across multiple data sets;
\citet{salzberg1997comparing} proposes Bonferroni-corrected ANOVA analysis.
\citet{demvsar2006statistical} reports that the Friedman non-parametric test with the Nemenyi correction makes fewer assumptions and has greater power than parametric tests.
Other authors \citep[e.g.,][]{luengo2009study,garcia2010advanced,derrac2011practical} further adapt the Friedman test for model comparison.

However, as \citet{demvsar2006statistical} notes, there still does not exist a non-parametric null hypothesis test designed for use with a repeated measure (i.e., $k$-fold) design across multiple data sets.
As a result, there is no procedure that takes into consideration the variance in scores of a given data set, at least within the frequentist paradigm.
\citet{demvsar2008appropriateness} and \citet{benavoli2017time}
enumerate additional problems with null hypothesis significance testing (NHST) procedures for model comparison:

\begin{itemize}
    \item NHST does not estimate probabilities for hypotheses; i.e., it does not tell us how likely two models are to perfrom equivalently,
    \item NHST $p$-values conflate effect size and sample size; i.e., with a sufficiently large sample, one can claim significance even if the effect size is trivial,
    \item NHST yields no information about the null hypothesis; i.e., one cannot draw further conclusions from a failure to reject the null hypothesis, and
    \item there is no principled way to select an appropriate $\alpha$-level for NHST.
\end{itemize}

\noindent
These issues lead \citeauthor{benavoli2017time} to reject NHST-based model comparison in favor of a Bayesian approach.
Bayesian hypothesis tests are defined by a likelihood function $p(d \mid \theta)$,
a probability model of the data $d$ conditioned on $\theta$, a vector of parameters.
The prior distribution for $\theta$, $p(\theta)$ must also be defined.
From these components, a posterior probability distribution $p(\theta \mid d)$ can then be calculated and queried (i.e., sampled from) to perform inference.
Various techniques can be used to estimate $\theta$;
they are usually related to the differences in models' scores using some evaluation metric, and ultimately, to whether one method is likely to perform better or worse than the other.
Thus, the posterior distribution can be used to perform model comparison.
\citeauthor{benavoli2017time} also introduce the notion of a \emph{region of practical equivalence} (henceforth, ROPE), which allows Bayesian hypothesis testing to estimate the likelihood that two models' results will be functionally indistinguishable. ROPE defines an interval around a model's result - if another model's performance falls within this interval - they are deemed practically equivalent. For example, if one deems that a difference of 1 percentage point in accuracy between models denotes practical equivalence, a $[-0.01, 0.01]$ interval is used as ROPE. If one model performs at $.941$ accuracy and another at $.949$ - they will be deemed practically equivalent. This allows to protect the statistical procedure from artifacts and false alarms of significance. Readers are referred to the accessible tutorial by \citet{benavoli2017time} for further details.

\citet{corani2017statistical} generalize Bayesian model comparison to a repeated measures scenario in which there are multiple data sets with unequal score variances.
They propose a hierarchical Bayesian model for estimating the likelihood of
one model performing better, worse, or equivalently, to another.
We now proceed to briefly describe and adapt this procedure to re-evaluate the findings of \citet{gorman-bedrick-2019-need}.

\section{Bayesian model comparison}
\label{sec:model-comparison}

Imagine a scenario where one wishes to compare the performance of two classifiers
across $q$ data sets.
By performing $m$ $k$-fold evaluations,
the experimenter obtains a vector of $n = mk$ observations,
i.e., differences in scores, between the two models:
$\mathbf{x}_{i} = (x_{i,1}, \dots, x_{i,n})$.
The values in these vectors are a positive cross-correlation $\rho$ because cross-validation introduces overlap in training data.
Let $\delta_i$ denote the mean difference score on the $i$th data set,
and let $\delta_0$ denote the average population-level difference.
\citet{corani2017statistical} propose a hierarchical probabilistic model

\begin{align}
\begin{aligned}
    \mathbf{x}_{i} & \sim \MVN(\mathbf{1} \delta_i, \Sigma_{i}), \\
    \delta_1 ... \delta_q & \sim t (\mu_0, \sigma_0, \nu),\\
    \sigma_1 ... \sigma_q & \sim \unif (0, \bar{\sigma})
\end{aligned}
\end{align}

\noindent
where $\MVN$ is a multivariate normal distribution over the vector of classifier 
differences with mean $\delta_i$ and 
a covariance matrix $\Sigma_{i}$ with variance $\sigma^2_{i}$ along the diagonal
and $\rho\sigma^2_{i}$ on the off-diagonal.
Data set variances are drawn from a Student distribution parameterized by the average population-level difference $\delta_0$ and variance $\sigma_0$,
with $\mu$ degrees of freedom.
The prior distributions for $\delta_0$, $\sigma_0$, and $\mu$ are defined so as to preserve the robustness of the model; these are motivated and described in more detail by \citet{corani2017statistical}.
Crucially, we model the differences obtained in individual runs using a multi-variate normal distribution oriented to the per-data set mean differences with a per-data set variance, and the mean differences using a unimodal distribution robust to outliers and non-normality.
Per-data set variances are modeled by a uniform distribution.

After the model learns the parameter distributions from experimental data, we obtain a posterior probability distribution $p(\delta_0, \sigma_0, \mu \mid d)$.
To infer whether one classifier is more likely to outperform another%
---or whether they are practically equivalent---%
we draw $N_{s}$ samples from the posterior distribution.
We use decision counters $n_{left}$, $n_{rope}$, and $n_{right}$ to keep track of how many times the left model was more likely to outperform the right model, 
to be practically equivalent to the right model,
and to be outperformed by the right model, respectively.
For each sample of the parameters, we define the posterior of the mean difference accuracy on a new unseen data set $\delta_{\text{next}}$ as $t(\delta_0, \sigma_0, \mu)$.
We obtain the outcome probabilities by integrating the distribution over the three intervals%
---e.g., we obtain the probability that the left model is better than the right by integrating from the left end of the distribution to the left edge of the ROPE interval,
and so on---%
and then incrementing the decision counter for the region with the highest outcome probability.
Finally, we compute likelihoods for the three scenarios by dividing the decision counts by the number of samples drawn:
$P(left) = \frac{n_{left}}{N_{s}}$,
$P(rope) = \frac{n_{rope}}{N_{s}}$, and
$P(right) = \frac{n_{right}}{N_{s}}$. 

Instead of significance, we thus estimate the likelihoods that one method is better than the other (or are practically equivalent).
These estimates follow from observing the beliefs of a Bayesian model that models the probability of the methods' mean difference on unseen data sets,
after sampling parameters from a meta-distribution which estimates the difference and variance over the population of data sets with $\mu$ degrees of freedom.

\section{Materials and methods}
\label{sec:methods}

To compare the results of our study with the ones obtained by \citet{gorman-bedrick-2019-need},
we use the same models,
data sets,
and evaluation metrics.%
\footnote{
    \url{http://github.com/kylebgorman/SOTA-taggers}
}
That is,
we compare implementations of 
the TnT \citep{TnT},
Collins \citep{Collins},
LAPOS \citep{LAPOS},
Stanford \citep{Stanford},
NLP4J \citep{NLP4J},
and Flair \citep{Flair}
part-of-speech taggers
using the Wall St.~Journal portions of the Penn Treebank (v.~3; \citealp{PTB})
and OntoNotes (v.~5; \citealp{OntoNotes}),
two widely-used corpora of American English financial news.
Summary statistics for this data are given in \cref{tab:datastats}.

\begin{table}
\centering
\begin{tabular}{l c r r}
\toprule
                && \# sentences & \# tokens \\
\midrule
{Penn Treebank} && 49,208 & 1,173,766 \\
{OntoNotes}     && 37,025 & 901,673 \\
\bottomrule
\end{tabular}
\caption{Summary statistics for the two corpora.}
\label{tab:datastats}
\end{table}

We perform 20 randomized 10-fold cross validations, obtaining 200 measurements of each model's performance on each data set.
In each run, 80\% of the data is used for training, 10\% for validation, and 10\% for evaluation.
We fit Bayesian models using the \texttt{baycomp} library%
\footnote{
    \url{http://github.com/janezd/baycomp}
}
and draw 50,000 samples from the posterior.

Following \citeauthor{gorman-bedrick-2019-need},
we use three evaluation metrics.
Token accuracy is simply the number of test data tokens correctly tagged divided by the total number of tokens, and is the standard intrinsic evaluation metric used for this task. 
OOV accuracy is similar to token accuracy but is restricted to out-of-vocabulary tokens,
i.e., those found in the test data but not in the training data.
Finally, sentence accuracy is the number of test data sentences that contain no tagging errors,
divided by the number of test sentences.
Ground-truth data is provided by human annotators.%
\footnote{
    Annotation quality for these data 
    has been studied by \citet{Ratnaparkhi1997}
    and \citet{Stanford}, among others.
}

\section{Results}
\label{sec:results}

Posterior distributions of the hierarchical models are visualized in \crefrange{fig:token.accuracy}{fig:oov.accuracy}
and summarized in \cref{tab:accuracy}.
We define the ROPE to have the same size as the 95\% confidence interval;
this is roughly 2--3\% for sentence and OOV accuracy, and 0.2\% for token accuracy.
Thus, two models are judged to be practically equivalent in sentence accuracy if they differ in performance on fewer than 98 sentences of the Penn Treebank or 75 sentences on the slightly smaller OntoNotes corpus.
For token accuracy, they are practically equivalent if they differ on fewer than 210 PTB tokens
or 162 OntoNotes tokens, respectively.

\begin{table*}
\centering
\begin{tabular}{ll c rrr c rrr c rrr}
\toprule
         &          && \multicolumn{3}{c}{Token accuracy} 
                    && \multicolumn{3}{c}{Sentence accuracy} 
                    && \multicolumn{3}{c}{OOV accuracy} \\
\cmidrule{4-6}
\cmidrule{8-10}
\cmidrule{12-14}
         &          && \multicolumn{1}{c}{$>$} & \multicolumn{1}{c}{$\approx$} & \multicolumn{1}{c}{$<$} 
                    && \multicolumn{1}{c}{$>$} & \multicolumn{1}{c}{$\approx$} & \multicolumn{1}{c}{$<$}
                    && \multicolumn{1}{c}{$>$} & \multicolumn{1}{c}{$\approx$} & \multicolumn{1}{c}{$<$} \\
\midrule
TnT      & Collins  &&  .168 &         .003  &  \textbf{.829} &&  .125 &         .001  &  \textbf{.874} &&  .158 & \textbf{.550} &         .292  \\
         & LAPOS    &&  .137 &         .001  &  \textbf{.862} &&  .127 &         .000  &  \textbf{.873} &&  .162 &         .305  & \textbf{.533} \\
         & Stanford &&  .109 &         .000  &  \textbf{.891} &&  .112 &         .000  &  \textbf{.888} &&  .088 &         .010  & \textbf{.903} \\
         & NLP4J    &&  .152 &         .000  &  \textbf{.848} &&  .156 &         .000  &  \textbf{.844} &&  .116 &         .002  & \textbf{.882} \\
         & Flair    &&  .158 &         .000  &  \textbf{.842} &&  .136 &         .000  &  \textbf{.864} &&  .094 &         .000  & \textbf{.906} \\
Collins  & LAPOS    &&  .116 & \textbf{.617} &          .267  &&  .105 &         .215  &  \textbf{.680} &&  .063 & \textbf{.842} &         .095  \\
         & Stanford &&  .180 & \textbf{.441} &          .379  &&  .124 &         .120  &  \textbf{.756} &&  .118 &         .038  & \textbf{.845} \\
         & NLP4J    &&  .137 &         .014  &  \textbf{.848} &&  .153 &         .010  &  \textbf{.837} &&  .166 &         .010  & \textbf{.824} \\
         & Flair    &&  .164 &         .000  &  \textbf{.836} &&  .157 &         .000  &  \textbf{.843} &&  .138 &         .001  & \textbf{.861} \\
LAPOS    & Stanford &&  .099 & \textbf{.822} &          .079  &&  .084 & \textbf{.829} &          .087  &&  .138 &         .091  & \textbf{.771} \\
         & NLP4J    &&  .172 &         .112  &  \textbf{.716} &&  .163 &         .137  &  \textbf{.700} &&  .161 &         .018  & \textbf{.821} \\
         & Flair    &&  .192 &         .004  &  \textbf{.805} &&  .190 &         .001  &  \textbf{.809} &&  .127 &         .003  & \textbf{.870} \\
Stanford & NLP4J    &&  .206 &         .122  &  \textbf{.672} &&  .191 &         .200  &  \textbf{.609} &&  .148 & \textbf{.441} &         .411  \\
         & Flair    &&  .197 &         .001  &  \textbf{.802} &&  .190 &         .001  &  \textbf{.809} &&  .130 &         .058  & \textbf{.812} \\
NLP4J    & Flair    &&  .150 &         .055  &  \textbf{.795} &&  .148 &         .024  &  \textbf{.827} &&  .092 & \textbf{.619} &         .288  \\
\bottomrule
\end{tabular}
\caption{Token, sentence, and OOV accuracy ranking likelihoods.}
\label{tab:accuracy}
\end{table*}

The hierarchical model estimates, for example, that TnT, the simplest tagger, would be outperformed in token accuracy by any of the other five taggers 80-90\% of the time.
However, there is a surprisingly high chance of practical equivalence in token accuracy between the Collins tagger, LAPOS, and the Stanford tagger;
for instance, the probability of practical equivalence of the latter two is 84\%.
This result is contrary to \citeauthor{gorman-bedrick-2019-need}'s replication of a standard split evaluation%
---they report that LAPOS is significantly better than the Collins tagger, and that the Stanford tagger is significantly better than LAPOS, according to two-tailed McNemar tests at $\alpha = .05$---%
but it is consistent with their subsequent failure to consistently reproduce this ranking using randomly-generated splits and Bonferroni-corrected McNemar tests.
In contrast, NLP4J and Flair are quite likely to outperform the other taggers, and Flair has an 80\% chance of outperforming NLP4J.

Similar results are obtained with sentence accuracy, a less-commonly used metric.
TnT is once again quite likely to be outperformed by other models.
Whereas LAPOS is quite likely to outperform the Collins tagger,
there is an 82\% probability that LAPOS and Stanford taggers will yield practically equivalent results.
Both NLP4J and Flair are both quite likely to outperform earlier models,
and Flair is most likely to outperform NLP4J.

There is a 55\% chance of practical equivalence between TnT and the Collins tagger for OOV accuracy.
This is somewhat surprising because the two models use rather different strategies for OOV inference:
TnT estimates hidden Markov model emission probabilities for OOVs using a simple suffix-based heuristic \citep[225f.]{TnT},
whereas the Collins tagger, a discriminatively-trained model, uses sub-word features developed by \citet{Ratnaparkhi1997} to handle rare or unseen words.
Similarly, whereas NLP4J and Flair also use distinct OOV modeling strategies,
we estimate that they have a 62\% likelihood to achieve practical equivalence on this metric.

\section{Discussion}
\label{sec:discussion}

Using the methods above, we obtain the following likelihood-based performance rankings:

\begin{itemize}
    \item token accuracy: TnT $<$  Collins $\approx$ LAPOS $\approx$ Stanford $<$ NLP4J $<$ Flair,
    \item sentence accuracy: TnT $<$ Collins $<$ LAPOS $\approx$ Stanford $<$ NLP4J $<$ Flair, and
    \item OOV accuracy: TnT $\approx$ Collins $\approx$ LAPOS $<$ Stanford $\lessapprox$ NLP4J $\approx$ Flair.
\end{itemize}
 
\noindent
We also find some divergences from the results reported by \citeauthor{gorman-bedrick-2019-need}.
For instance, they find that the Stanford tagger has significantly higher token accuracy than LAPOS on the Penn Treebank standard split.
According to our model, the two taggers are most likely practically equivalent,
a result which is consistent with their later finding that Stanford outperforms LAPOS on only 1 out of 20 Penn Treebank random splits. We also find out that while both taggers were practically equivalent in both token and sentence accuracy, Stanford is likely to outperform LAPOS in OOV words, which could have impacted the statistical significance in the original experiment, as the repetition of the k-fold procedure causes strong variation - of the vocabulary available at training and OOV token sets - between experimental runs. 

We note that Bayesian comparison and the precise quantities it estimate may give insights into the particular strengths and weaknesses of the various models and evaluation metrics.
For instance, we infer that whereas
the Collins tagger improves upon TnT,
and Flair improves upon NLP4J, in both token and sentence accuracy,
these improvements are not likely to be due to differences in the models' handling of out-of-vocabulary words.
This is because TnT and the Collins tagger,
and NLP4J and Flair, are most likely practically equivalent in their tagging accuracy for OOV words.

In Bayesian approaches as we are thinking about probabilities of a method outperforming another method. As a result we can do what was not possible in the NHST approach taken by Gorman and Bedrick. We can order methods into at a partial ordering to gain an insight into which methods are more likely to perform better than others. We can do this based on the modeled likelihoods, but it would not be in a NHST framework, because there are currently no multiple comparison correction procedures that take into account the variance of repeated runs of a method on the same data set. 

Gorman and Bedrick reported that LAPOS would be sure to outperform Collins on PTB (20 out of 20 times), but not on Ontonotes (7 out of 20 times) in token accuracy. We found out that that the most likely scenario, when the performance is modeled using a hierarchical model on evidence from both data sets jointly, that these difference are likely within practical equivalence. 

We set the interval of practical equivalence of observed accuracies to match the 95\% confidence intervals reported by Bedrick and Gorman, to maintain a capacity for comparing the two experimental approaches. However, we believe it is much more useful to have an interpretable and intuitively understandable definition of what practical equivalence means in the experiment. Instead of setting it based on statistical confidence intervals, we recommend selecting the ROPE to represent the scale of human annotator differences, or the error level that does not negatively impact a downstream task that depends on the prediction quality of evaluated methods.

\section{Conclusions}
\label{sec:conclusions}

We compare the performance of six part-of-speech taggers on two data sets
using twenty repetitions of a ten-fold cross-validation procedure
and statistical system comparison performed using hierarchical Bayesian models.
By sampling from the posterior distribution of these models,
we estimate the likelihood that
a given tagger will be better than, worse than, or practically equivalent to other taggers
on three different evaluation metrics.
These estimates are valid insofar as the data sets used to estimate the Bayesian models comprise a representative sample of a coherent population of data sets.
This method provides a principled way to perform statistical model comparison using $k$-fold cross-validation, a data-efficient evaluation technique.
It also allows us to incorporate results obtained across multiple data sets and to make population-level inferences.
We finally compare the results obtained with the proposed method to those computed using randomly generated splits and traditional NHST-based model comparison.
The results provide new insights into the strengths and weaknesses of English part-of-speech tagging models, complementing other approaches to model comparison and interpretation.

\section*{Acknowledgments}
We would like to thank Steve Bedrick for previous work on this topic. This work was supported by the statutory funds of the Department of Computational Intelligence, Wroclaw University of Science and Technology.

\bibliographystyle{acl_natbib}
\bibliography{bayesian}
\newpage
\appendix
\section{Visualizations}
Visualizations of the posterior samples in \cref{sec:results} are shown 
in \crefrange{fig:token.accuracy}{fig:oov.accuracy} below.

\label{sec:appendix}
\begin{figure*}[!h]
\begin{tcbraster}[raster columns=3, enhanced, blankest, tile]
	\tcbincludegraphics{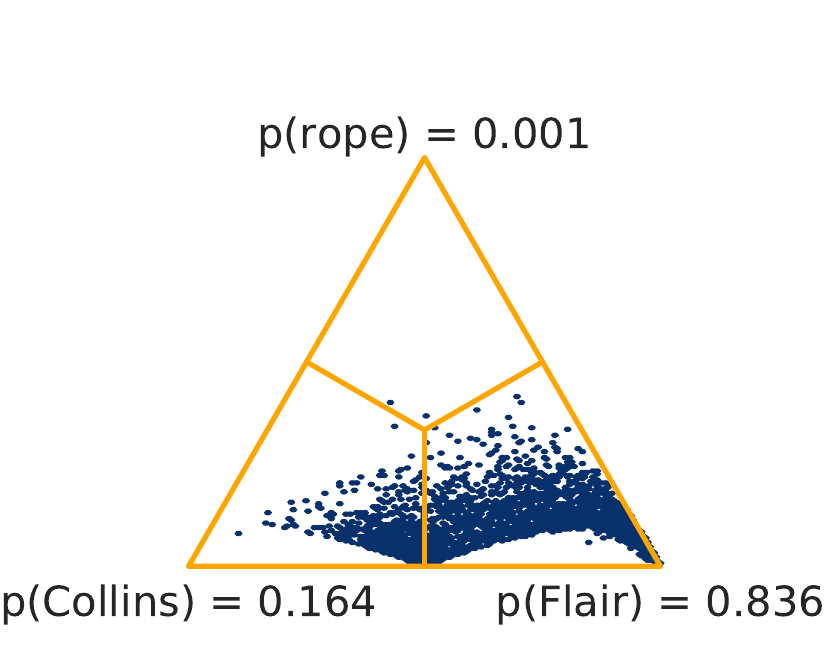}
	\tcbincludegraphics{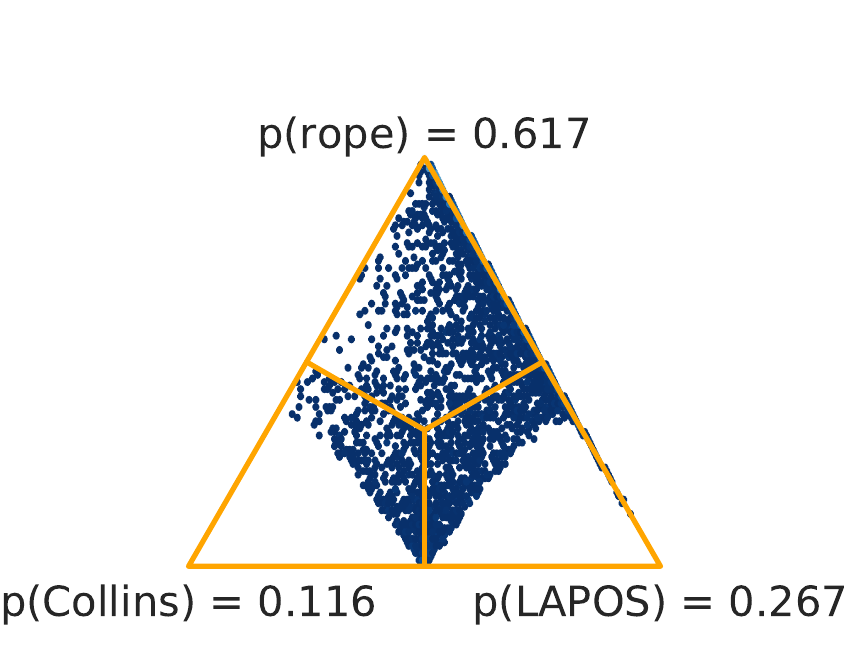}
	\tcbincludegraphics{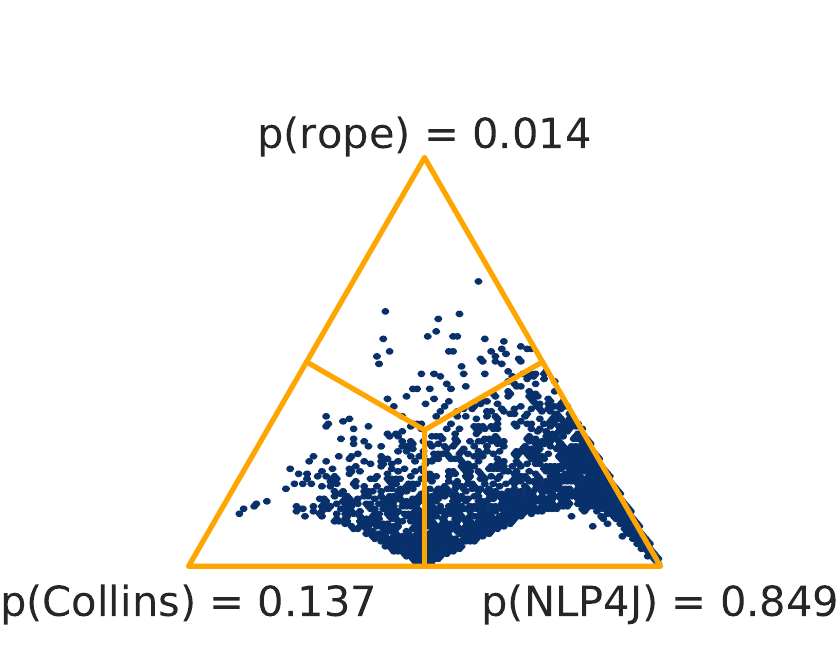}
	\tcbincludegraphics{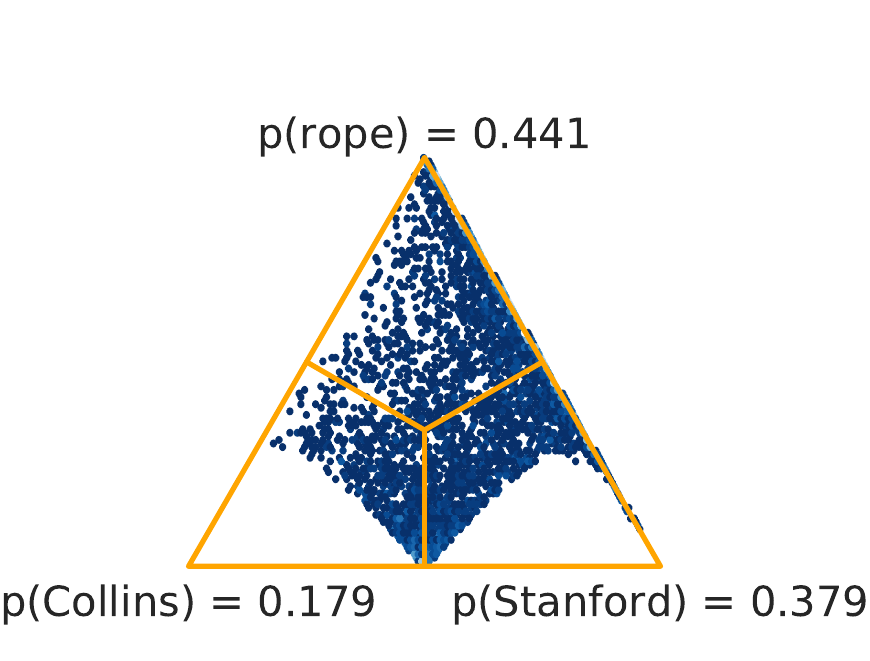}
	\tcbincludegraphics{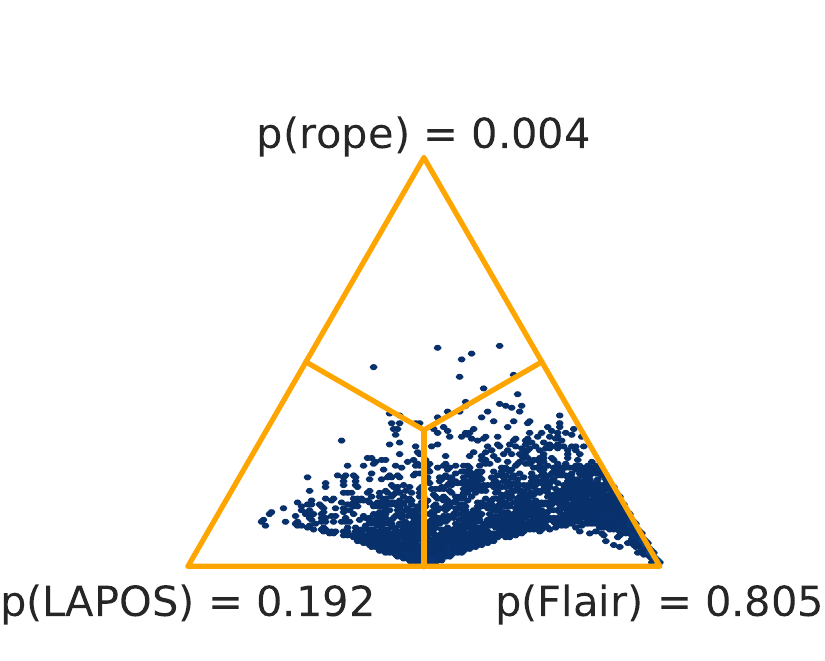}
	\tcbincludegraphics{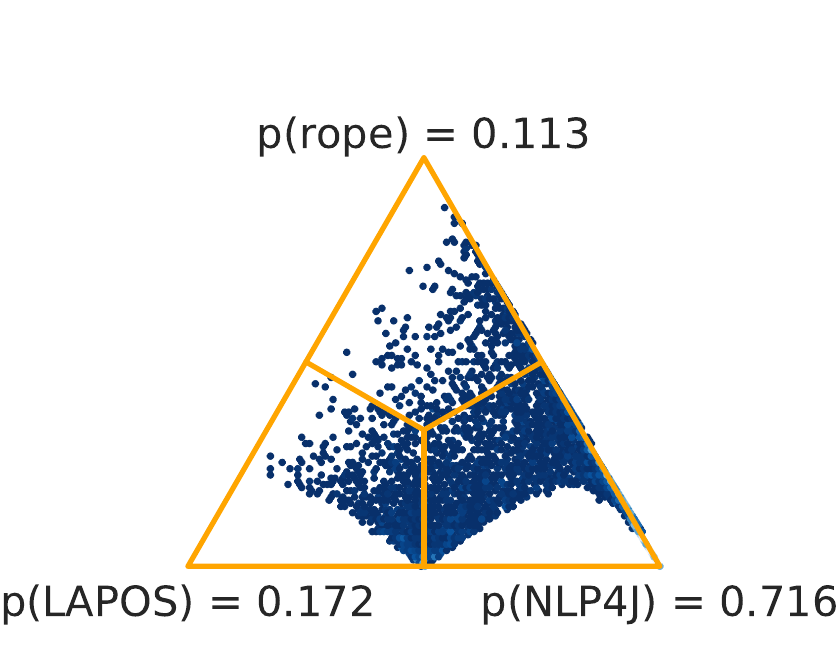}
	\tcbincludegraphics{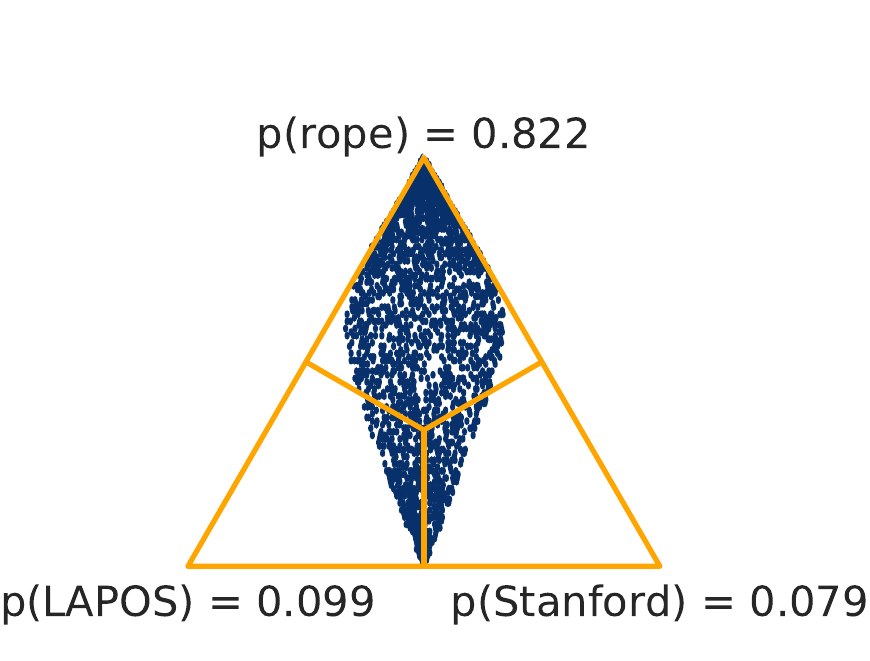}
	\tcbincludegraphics{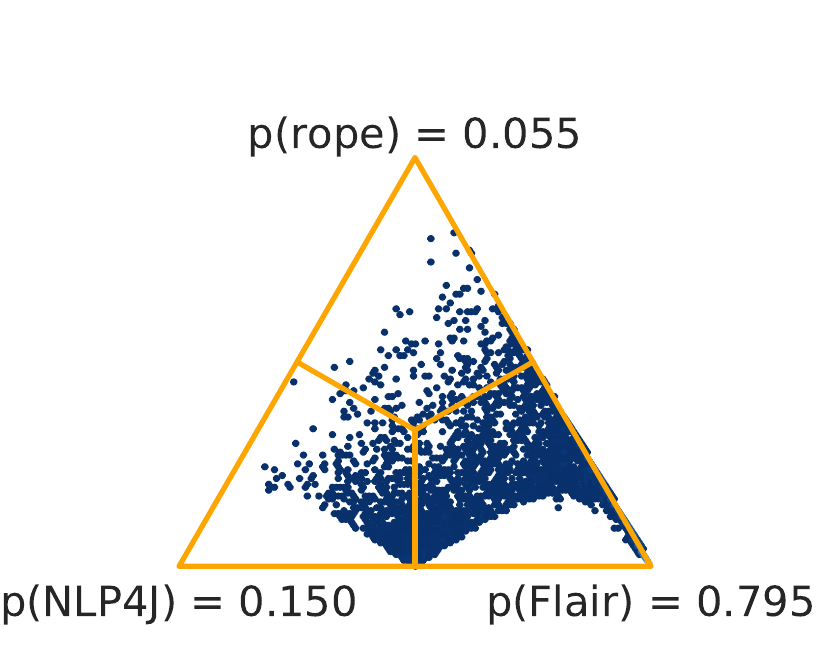}
	\tcbincludegraphics{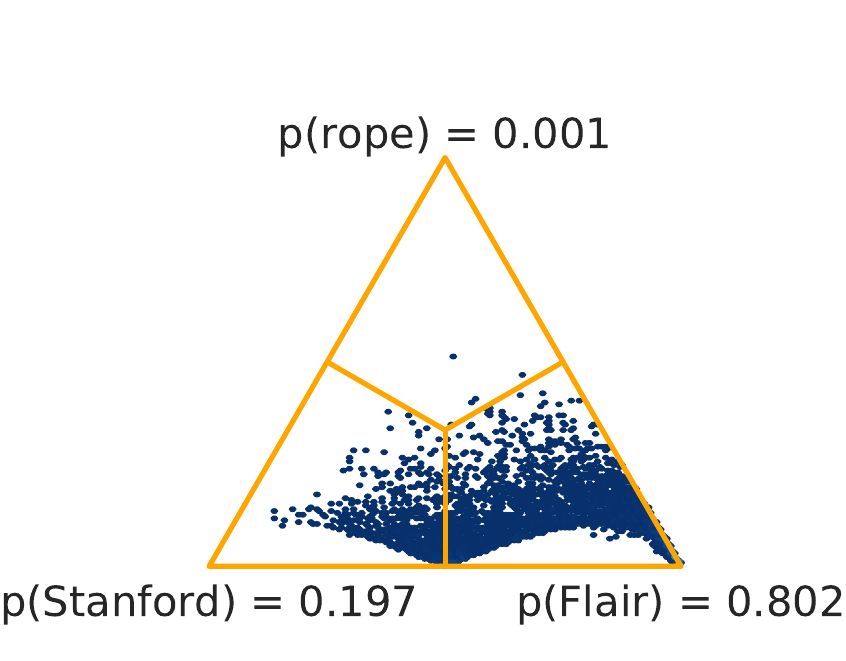}
	\tcbincludegraphics{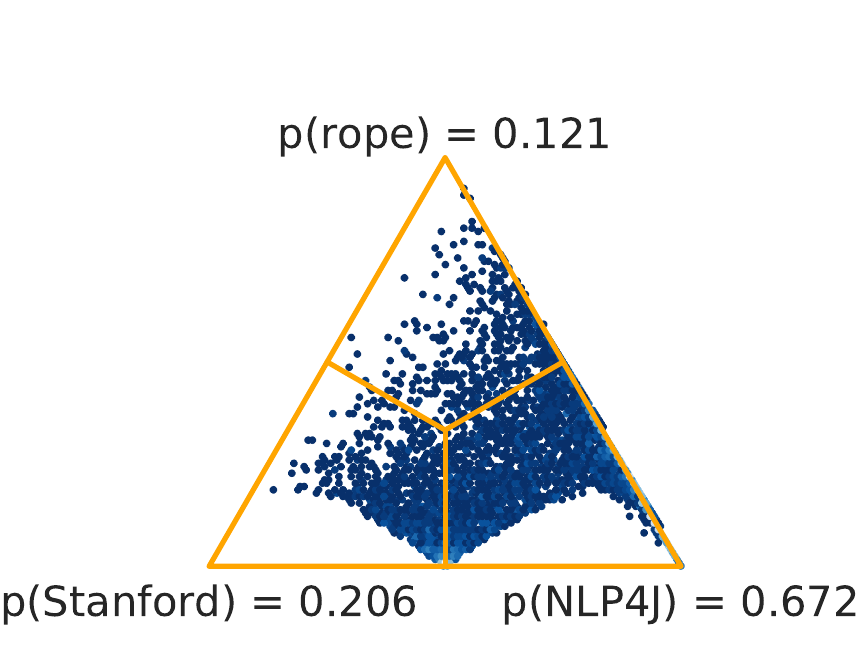}
	\tcbincludegraphics{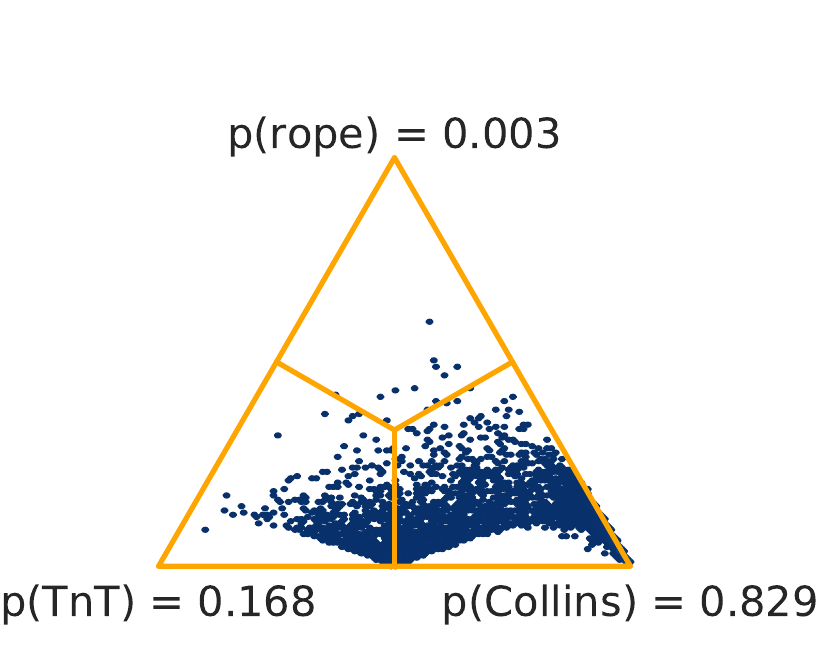}
	\tcbincludegraphics{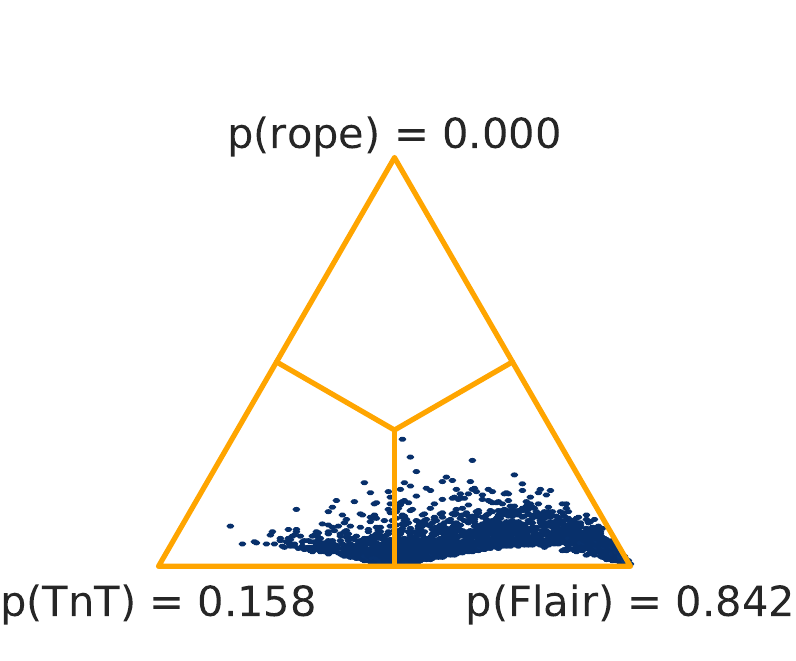}
	\tcbincludegraphics{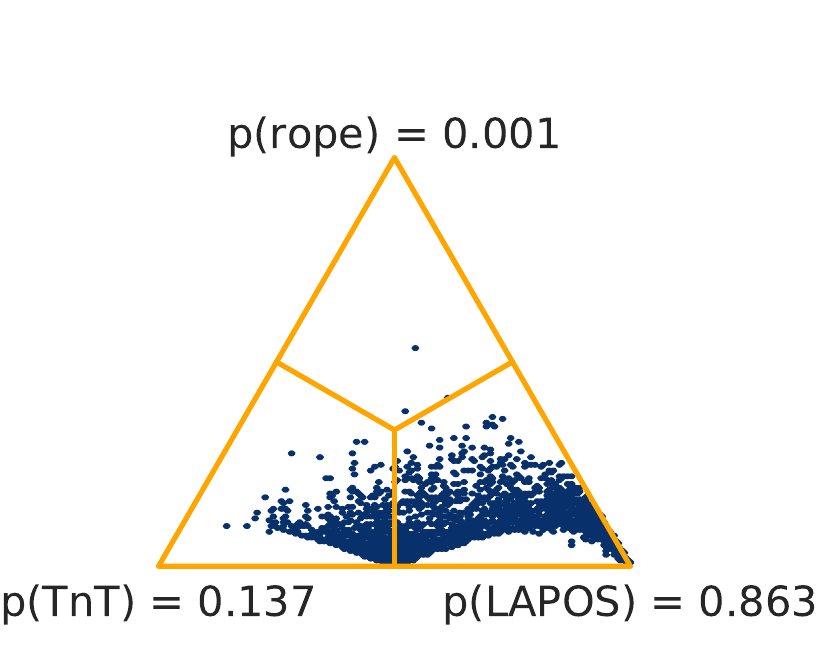}
	\tcbincludegraphics{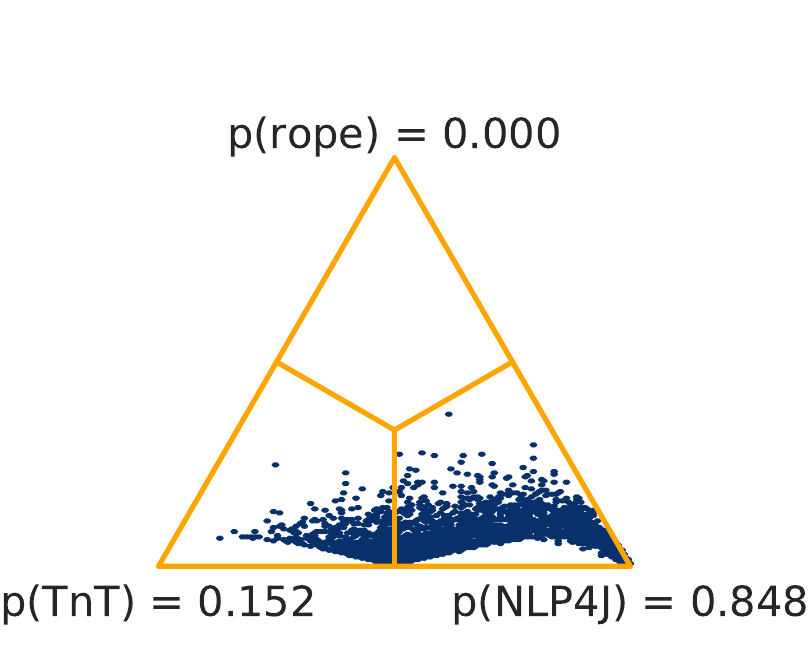}
	\tcbincludegraphics{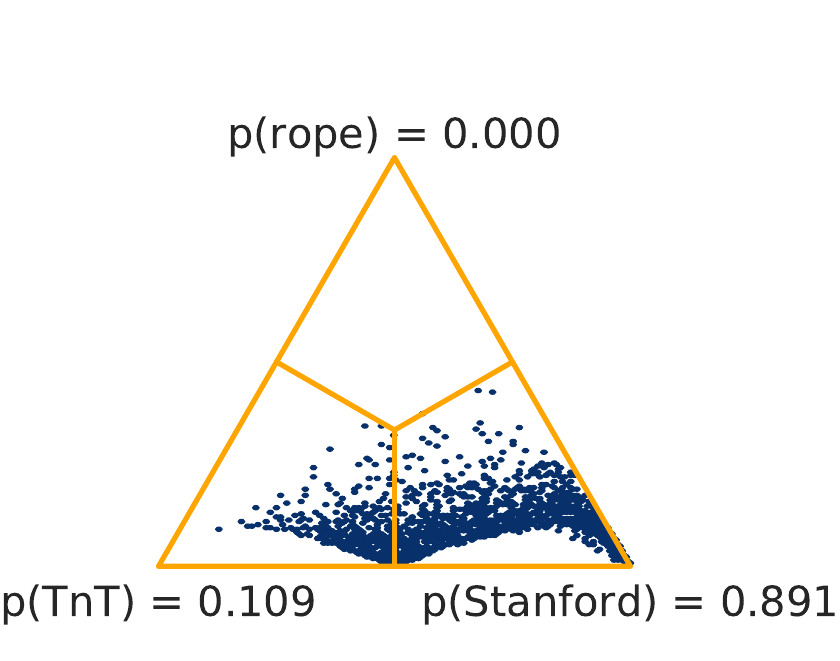}
\end{tcbraster}
\caption{Pairwise comparisons of models' token accuracy; the triangles illustrate 50,000 samples drawn from the posterior distribution, and the likelihood that a given method would perform better, or that their results would be practically equivalent.}
\label{fig:token.accuracy}
\end{figure*}

\newpage
\begin{figure*}
\begin{tcbraster}[raster columns=3, enhanced, blankest]
	\tcbincludegraphics{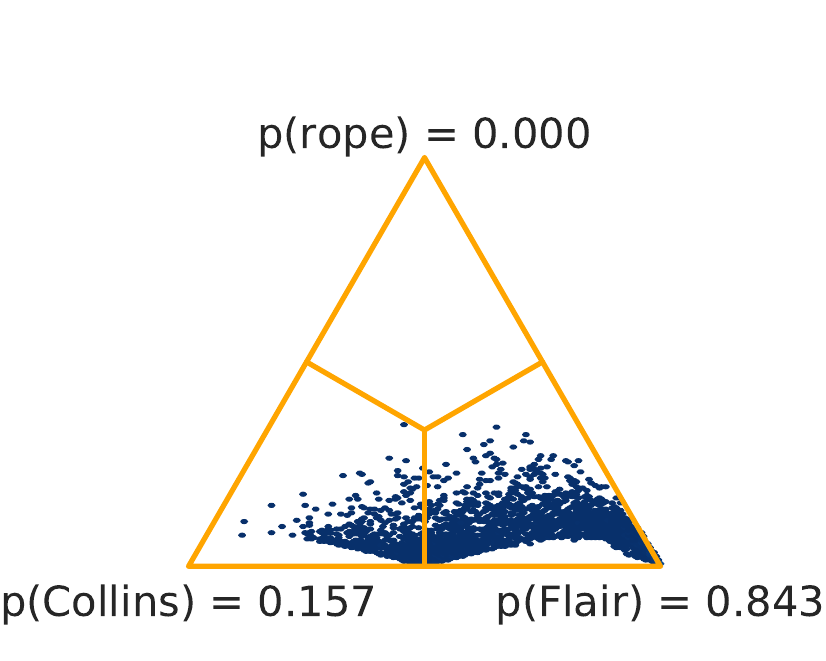}
	\tcbincludegraphics{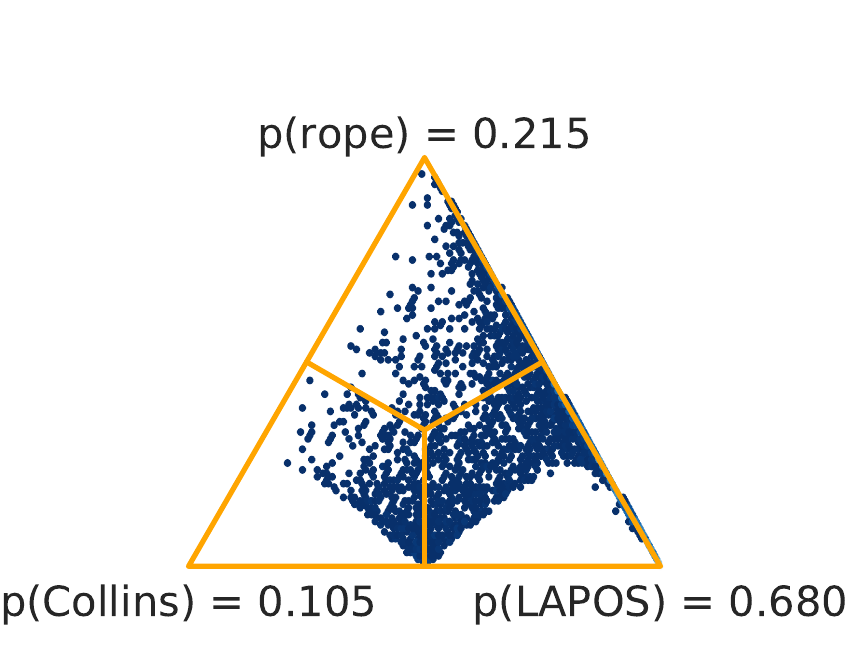}
	\tcbincludegraphics{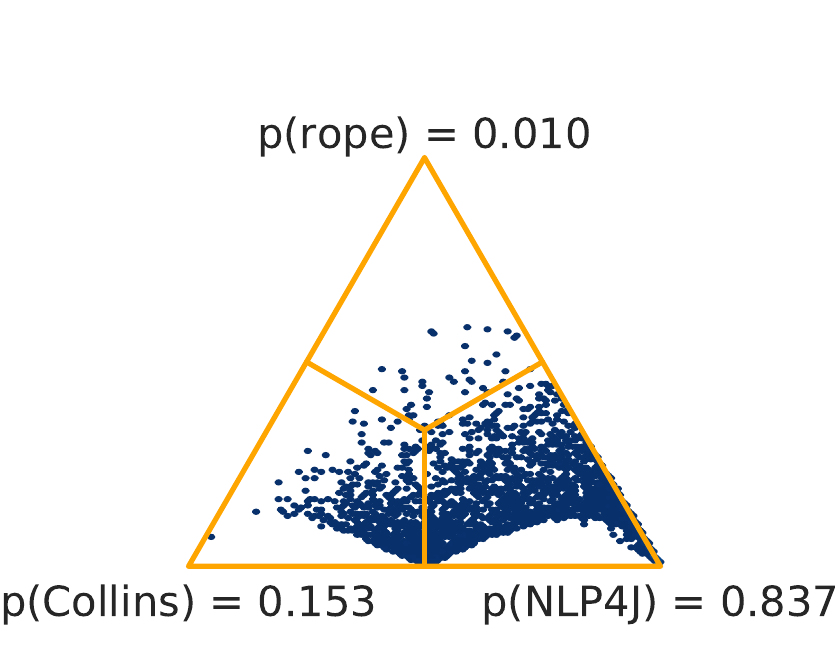}
	\tcbincludegraphics{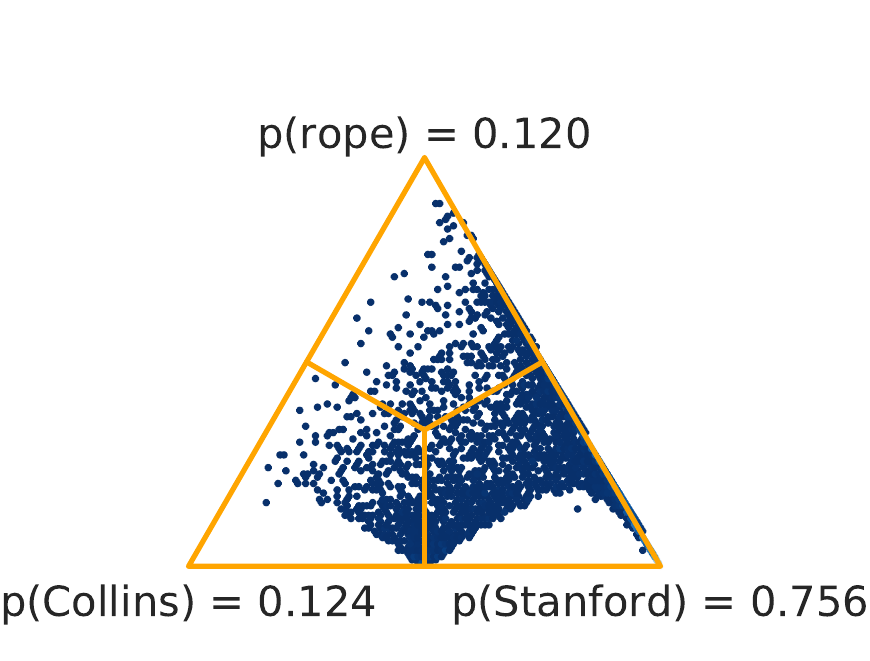}
	\tcbincludegraphics{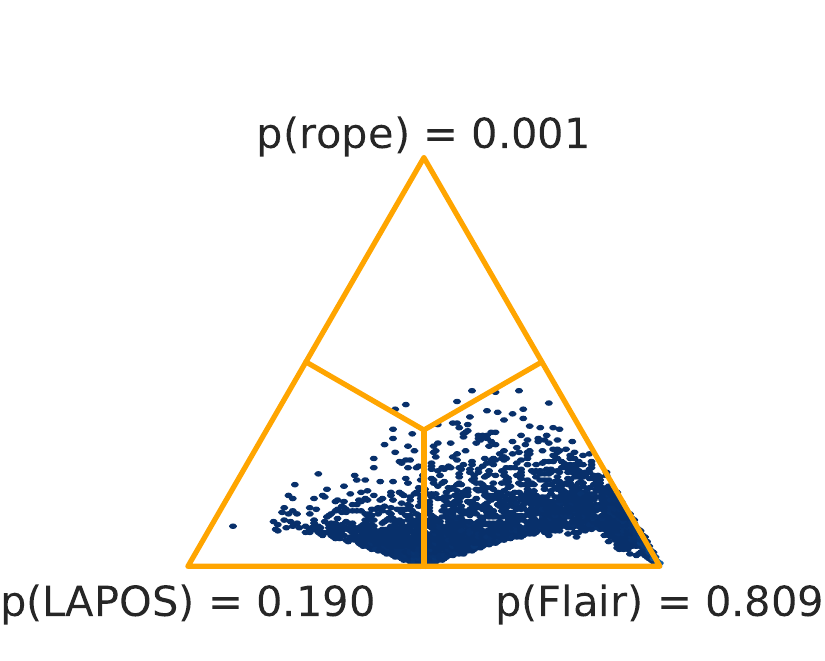}
	\tcbincludegraphics{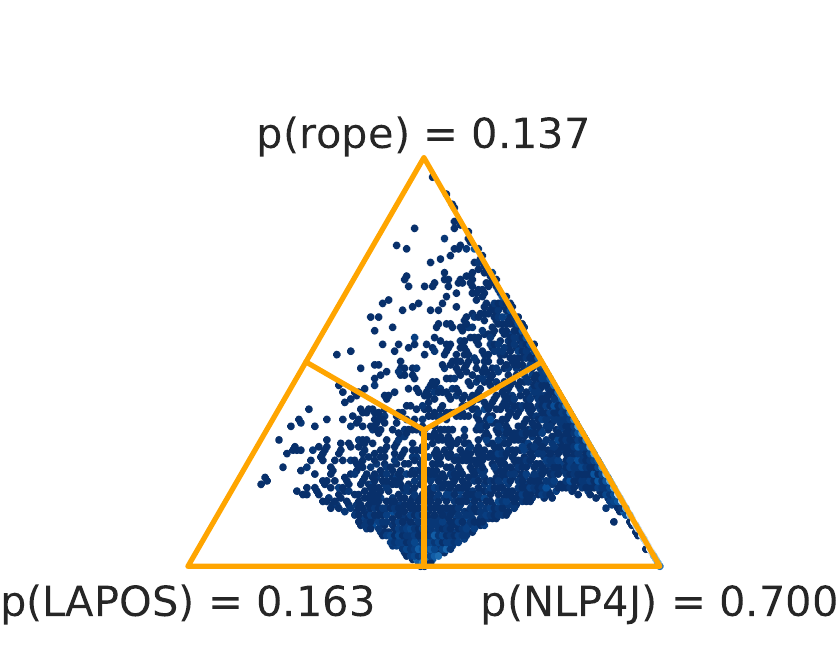}
	\tcbincludegraphics{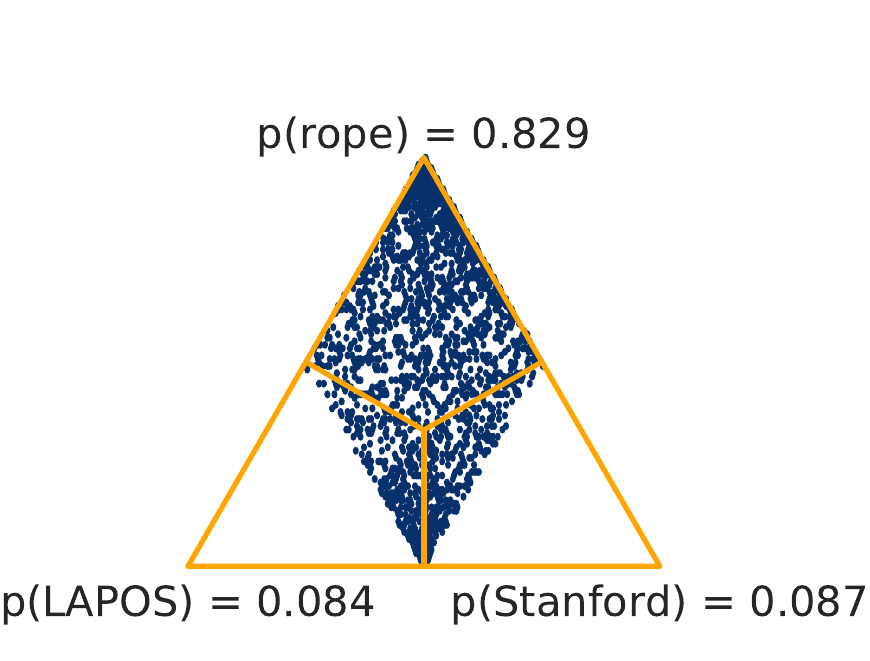}
	\tcbincludegraphics{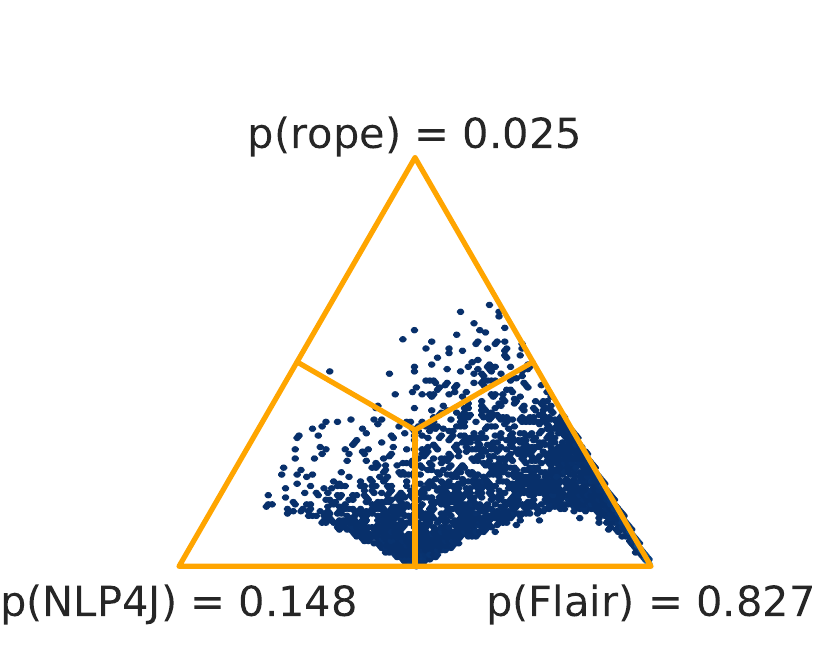}
	\tcbincludegraphics{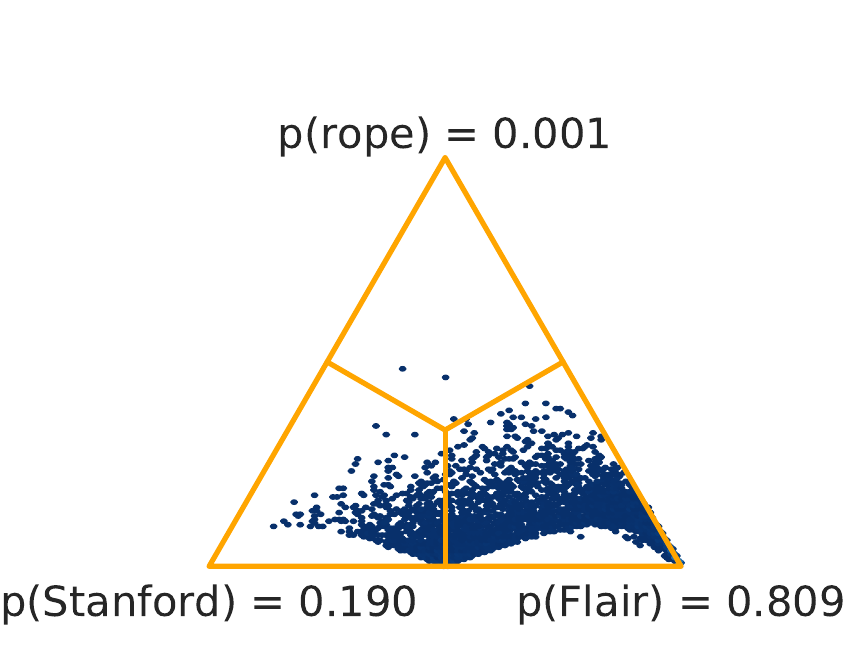}
	\tcbincludegraphics{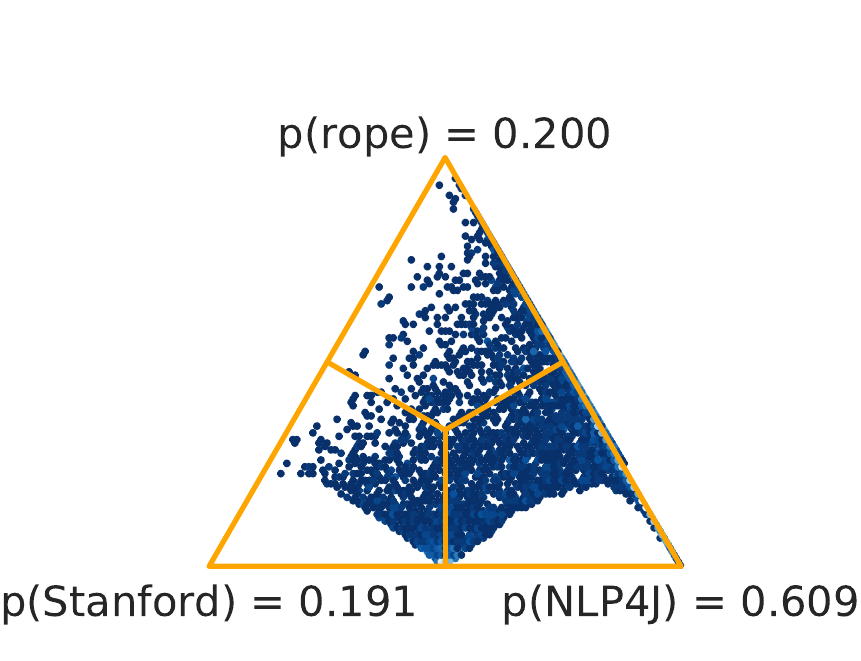}
	\tcbincludegraphics{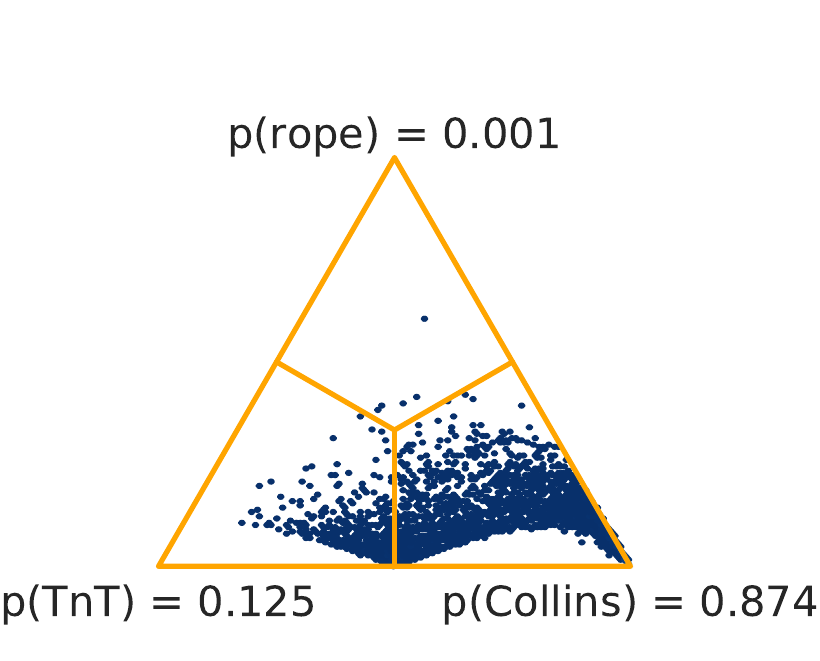}
	\tcbincludegraphics{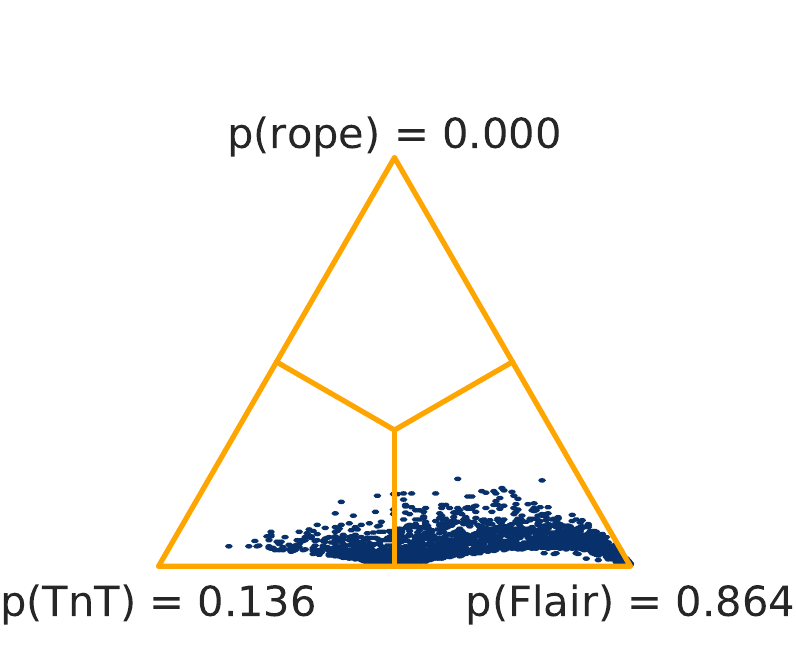}
	\tcbincludegraphics{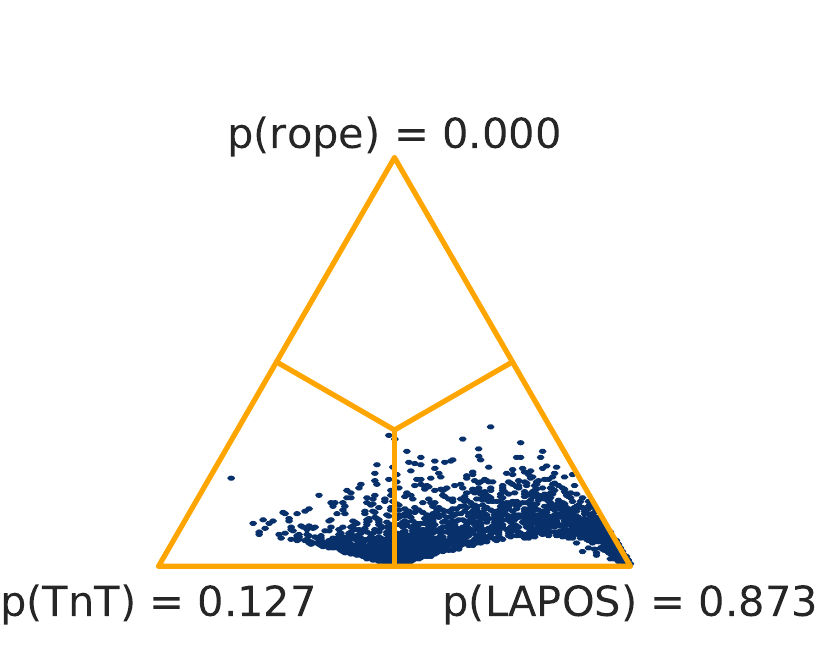}
	\tcbincludegraphics{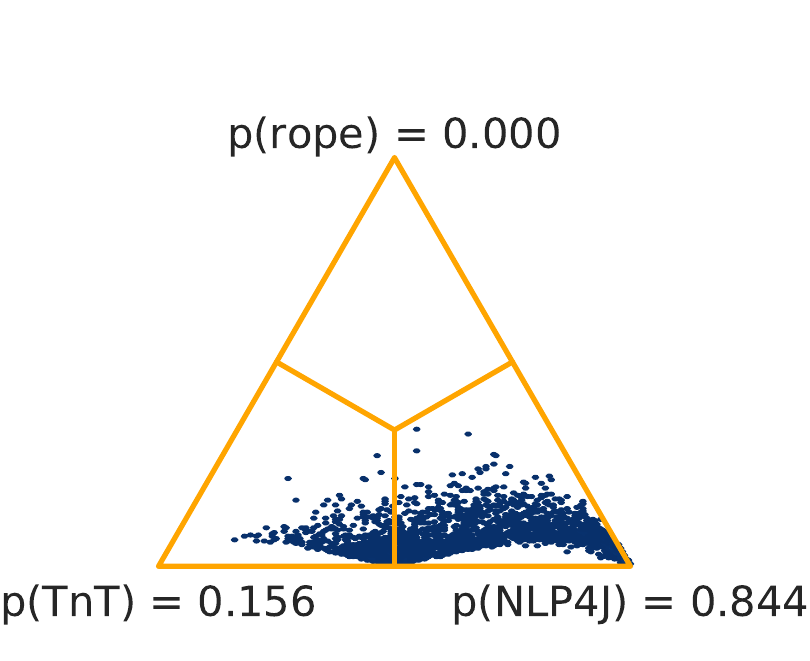}
	\tcbincludegraphics{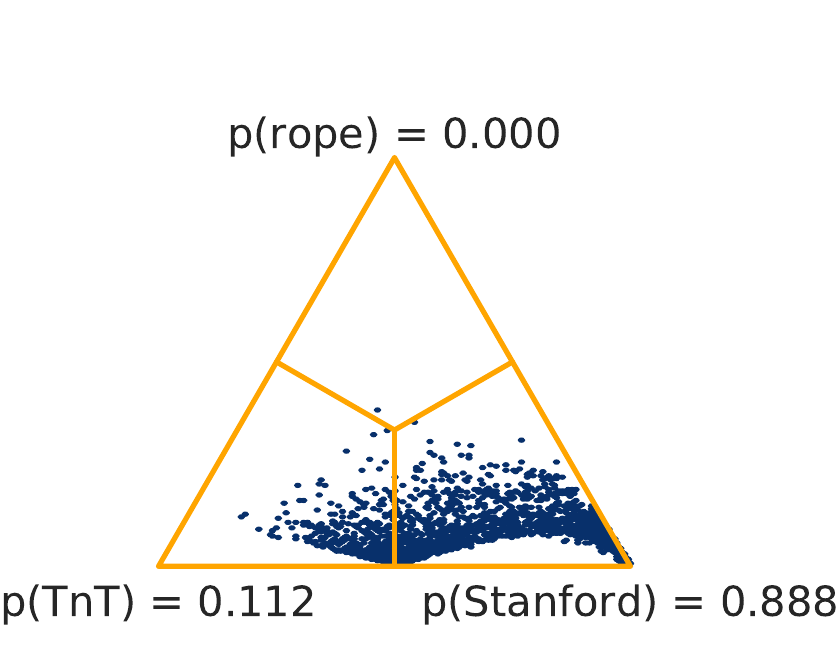}
\end{tcbraster}
\caption{Pairwise comparisons of models' sentence accuracy; the triangles illustrate 50,000 samples drawn from the posterior distribution alongside the likelihood that a given method would perform better, or that their results would be practically equivalent.}
\label{fig:sentence.accuracy}
\end{figure*}

\newpage
\begin{figure*}
\begin{tcbraster}[raster columns=3, enhanced, blankest]
	\tcbincludegraphics{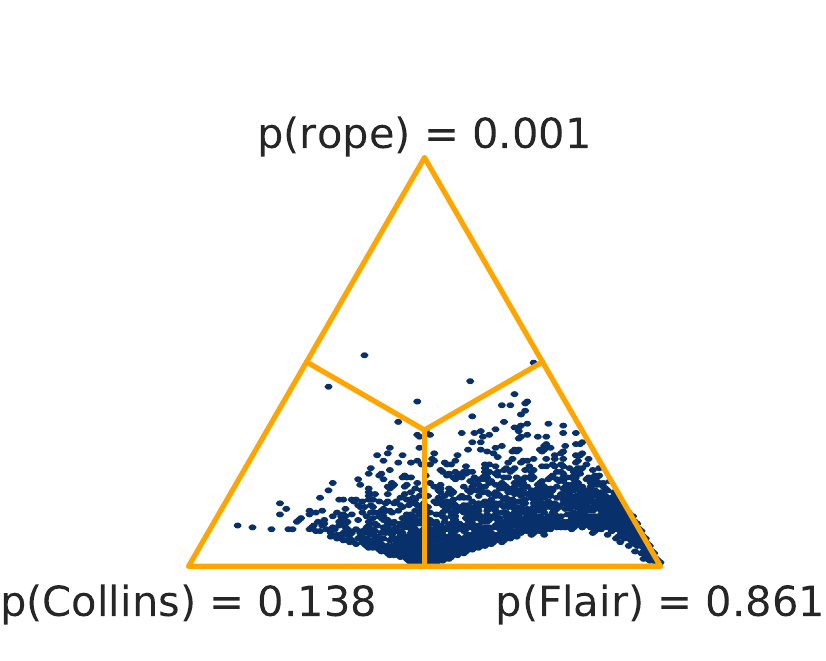}
	\tcbincludegraphics{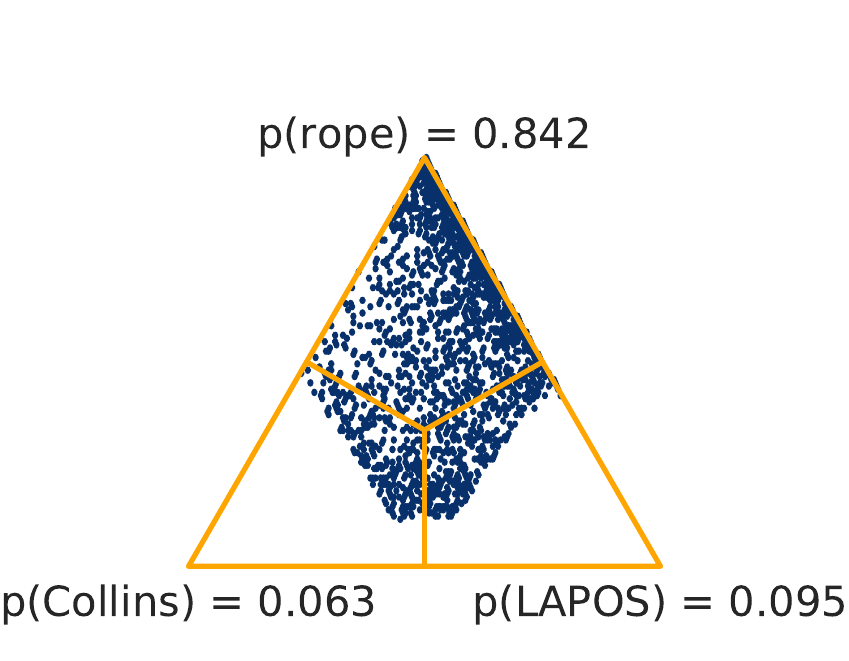}
	\tcbincludegraphics{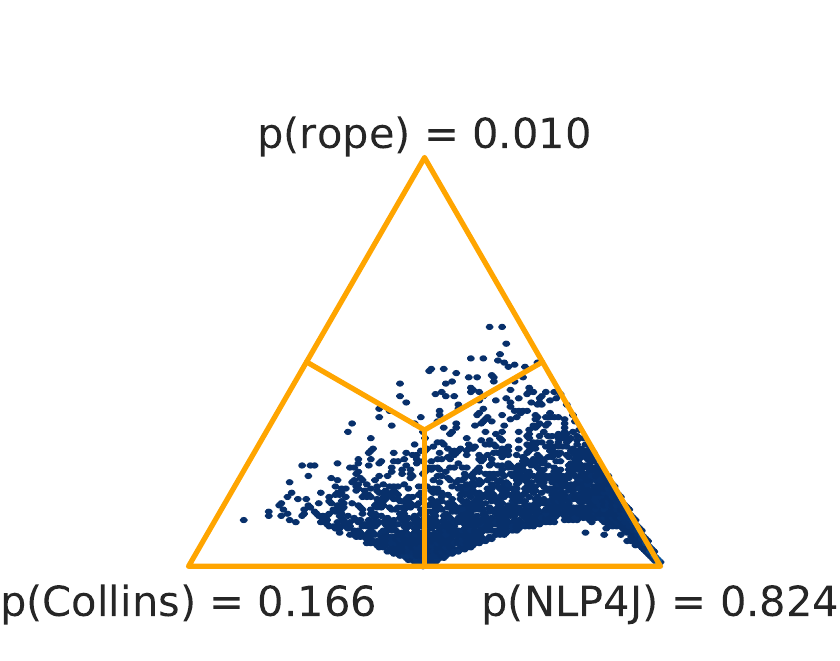}
	\tcbincludegraphics{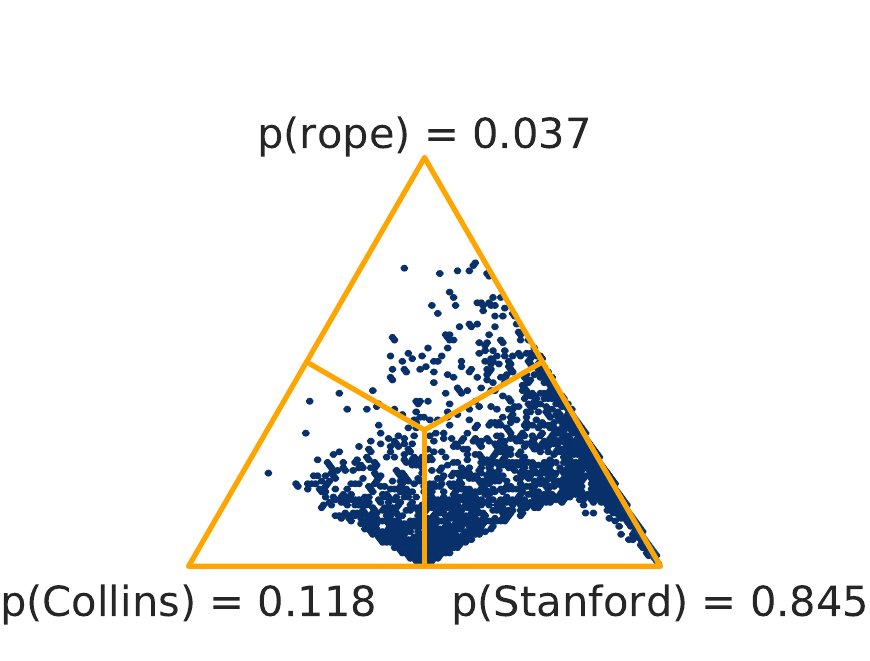}
	\tcbincludegraphics{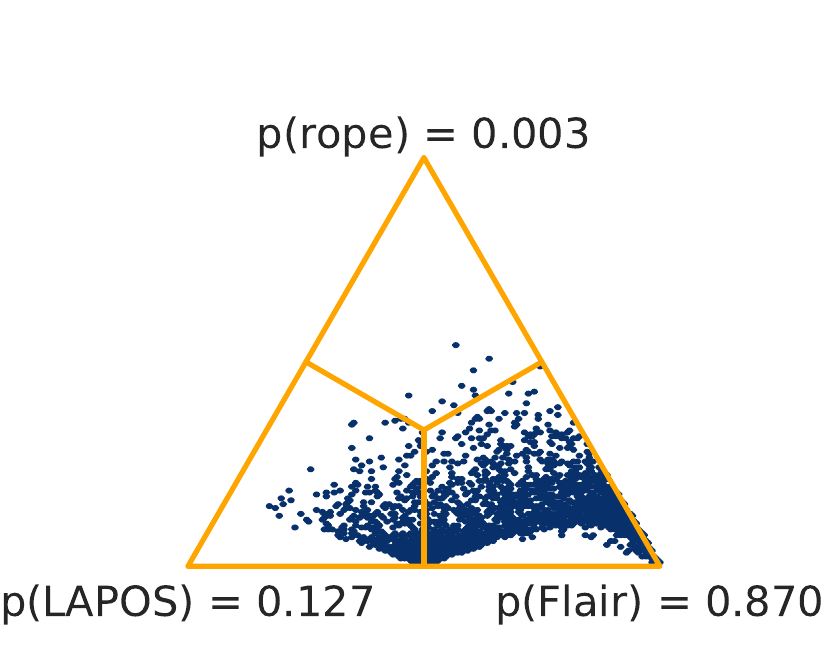}
	\tcbincludegraphics{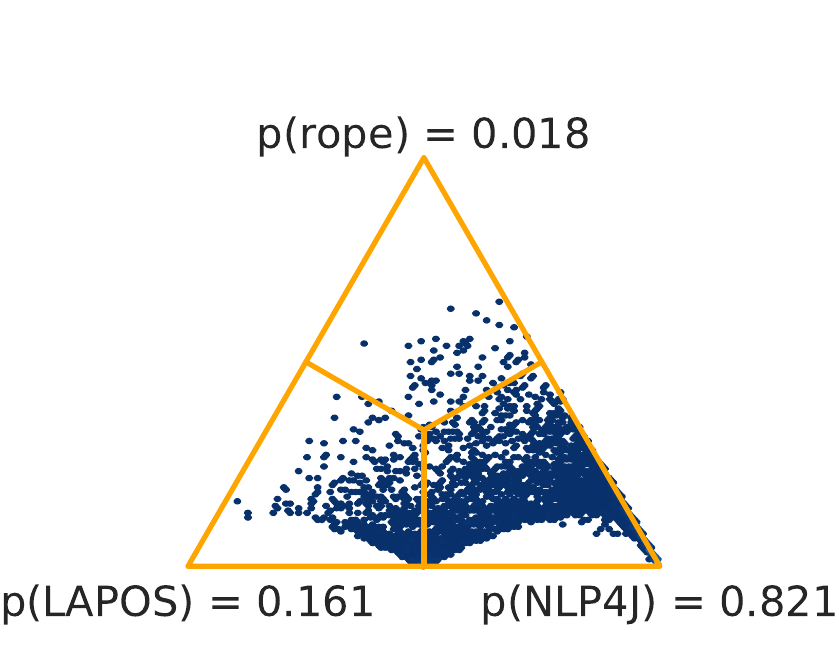}
	\tcbincludegraphics{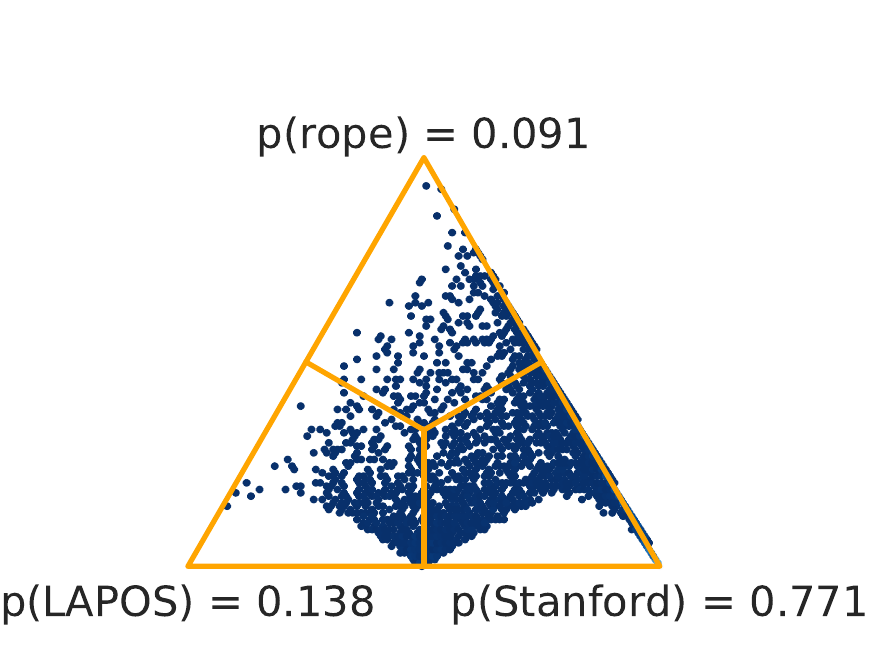}
	\tcbincludegraphics{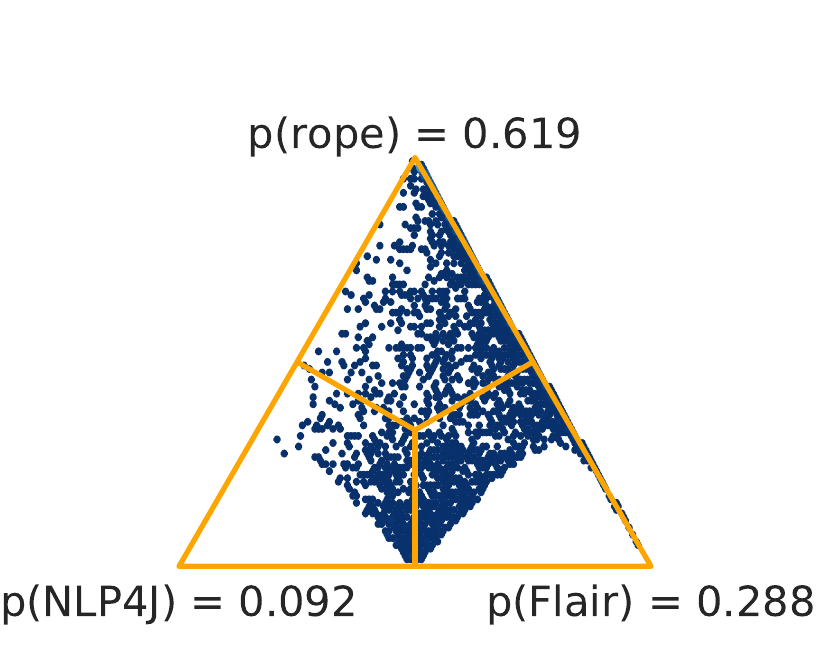}
	\tcbincludegraphics{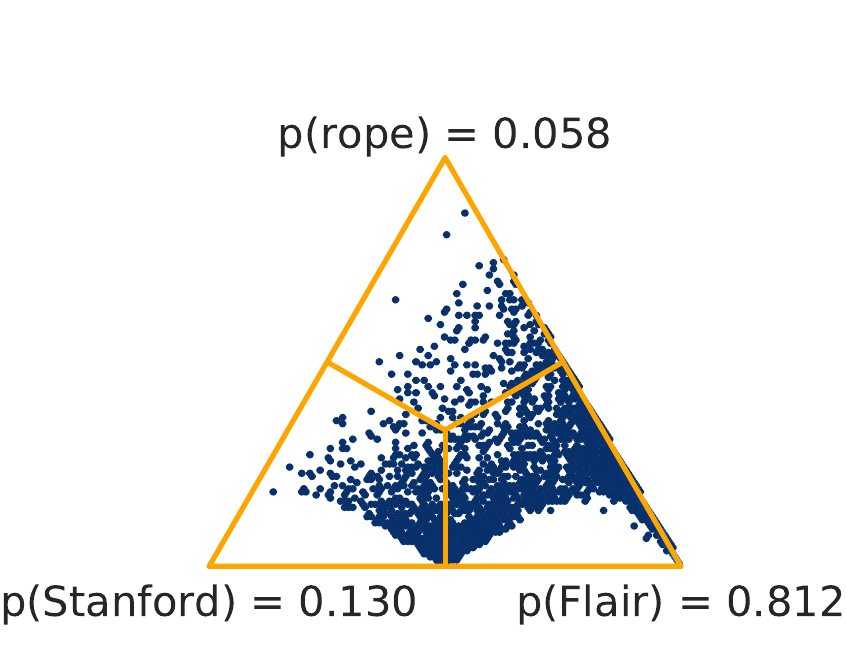}
	\tcbincludegraphics{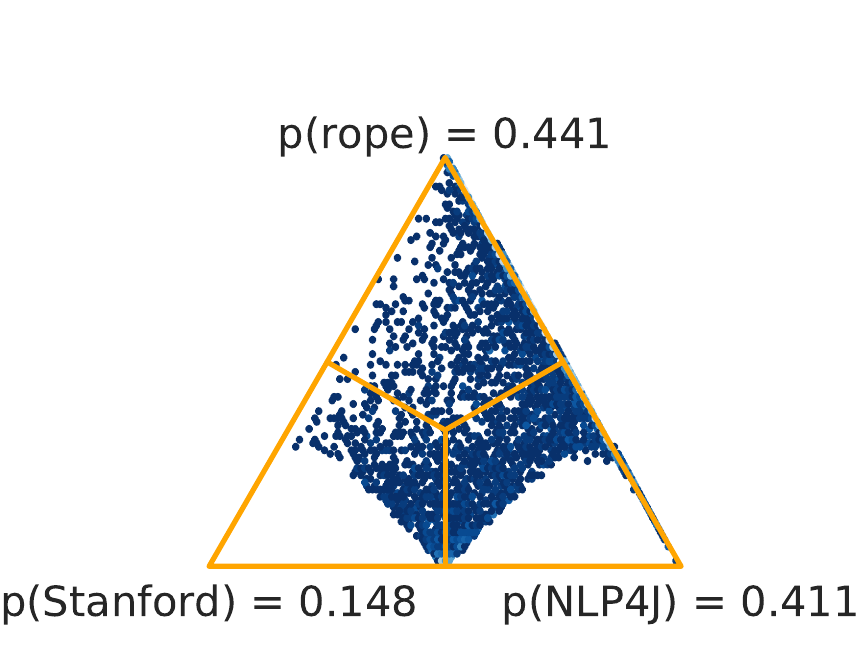}
	\tcbincludegraphics{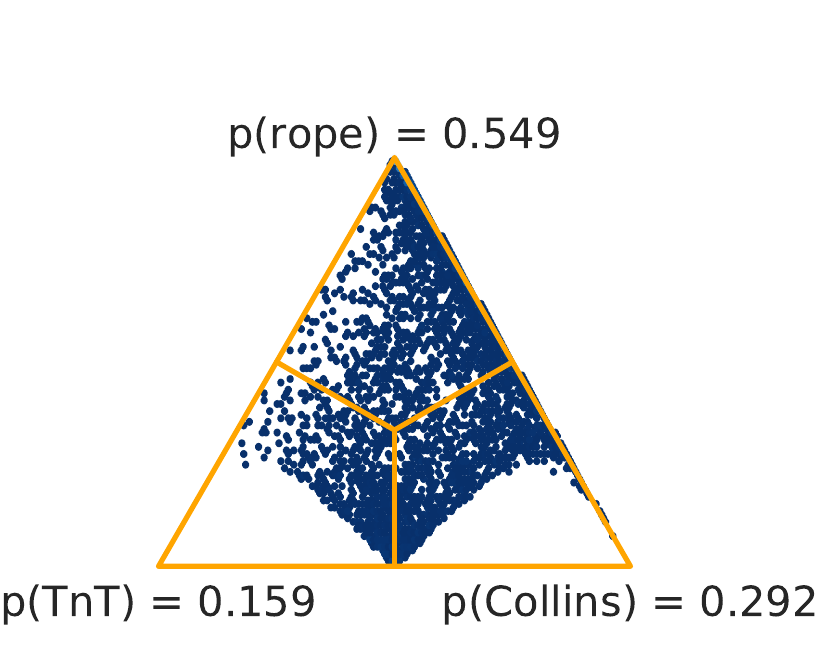}
	\tcbincludegraphics{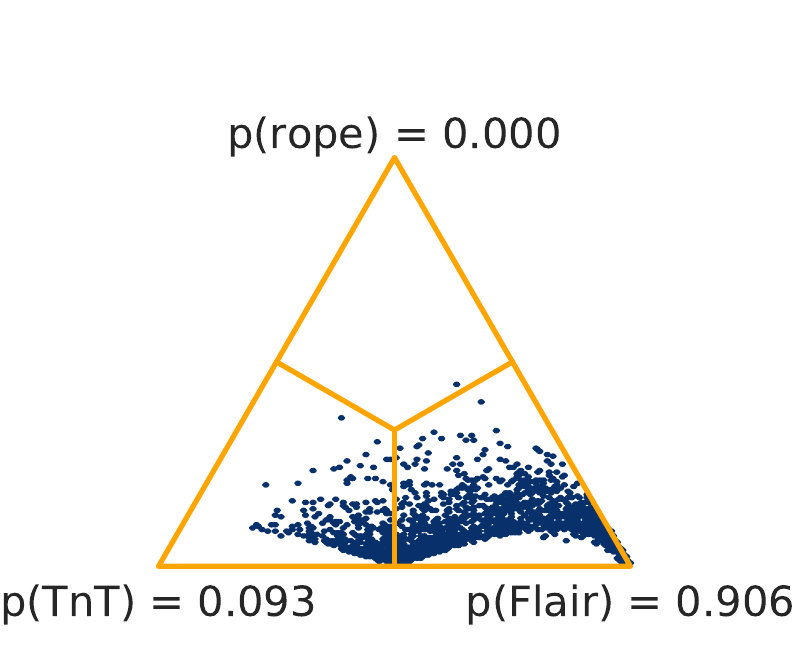}
	\tcbincludegraphics{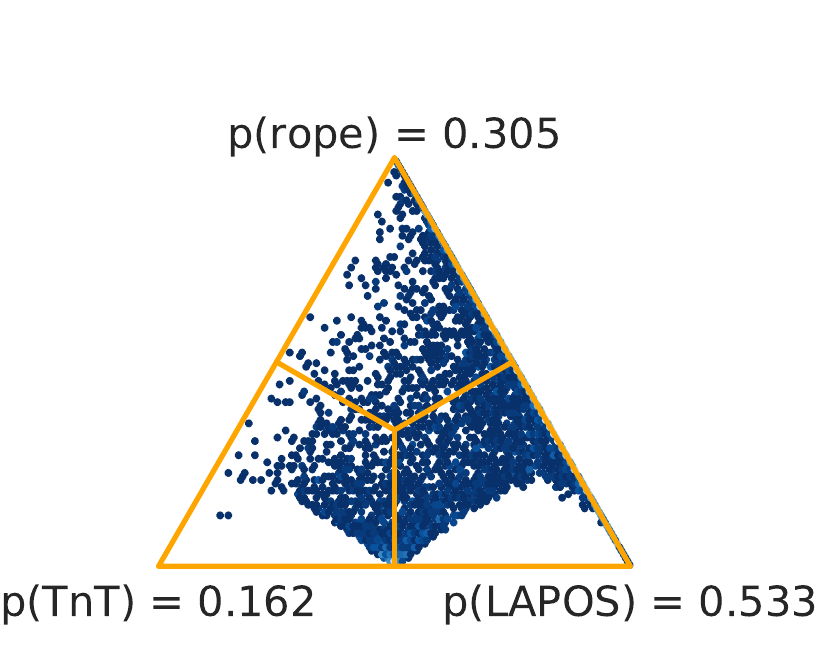}
	\tcbincludegraphics{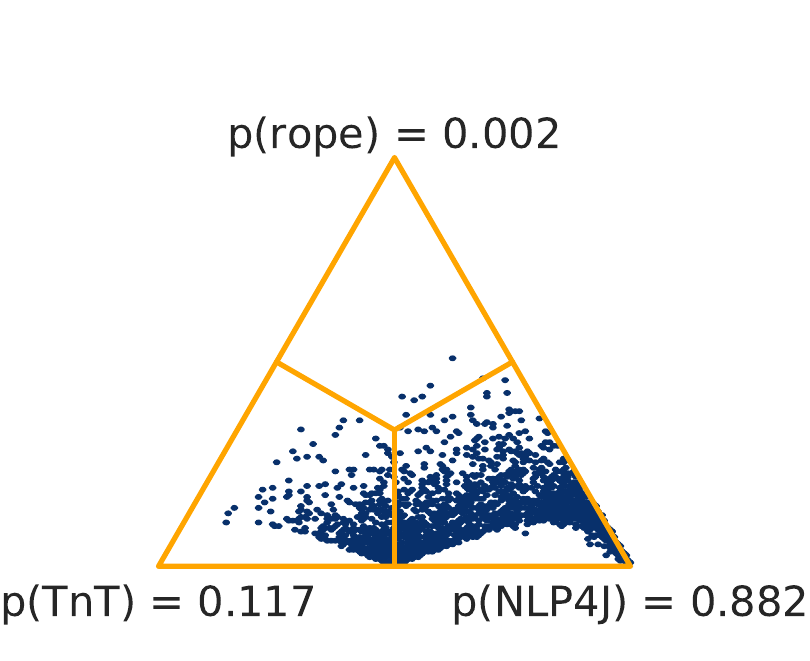}
	\tcbincludegraphics{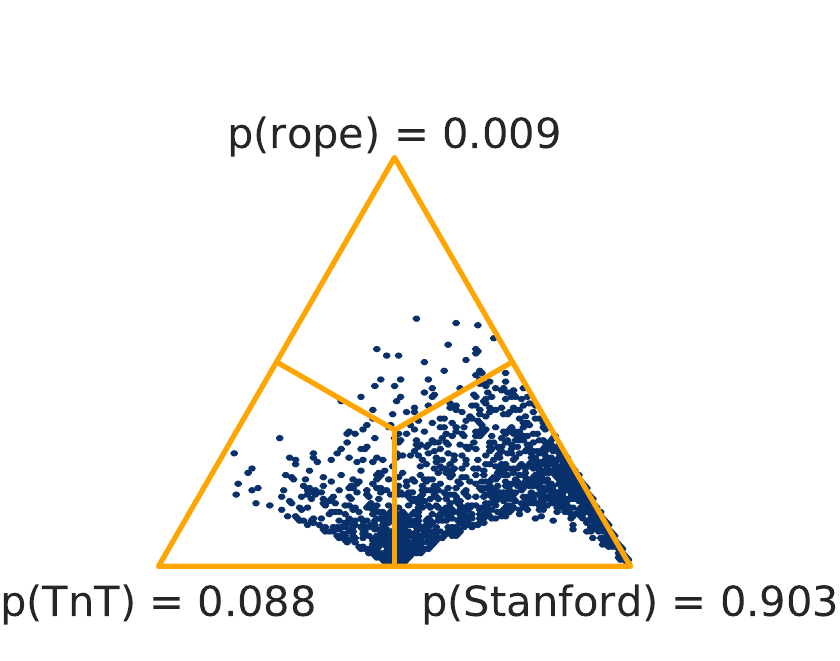}
\end{tcbraster}
\caption{Pairwise comparisons of models' OOV accuracy; the triangles illustrate 50,000 samples drawn from the posterior distribution alongside the likelihood that a given method would perform better, or that their results would be practically equivalent.}
\label{fig:oov.accuracy}
\end{figure*}

\end{document}